\documentclass[conference]{IEEEtran}
\IEEEoverridecommandlockouts
\usepackage{cite}
\usepackage{amsmath,amssymb,amsfonts}
\usepackage{algorithmic}
\usepackage{graphicx}
\usepackage{textcomp}
\usepackage{xcolor}
\usepackage{times}
\usepackage{soul}
\usepackage{url}
\usepackage{bm}
\usepackage[hidelinks]{hyperref}
\usepackage[utf8]{inputenc}
\usepackage[small]{caption}
\usepackage{graphicx}
\usepackage{booktabs}
\usepackage{helvet}
\usepackage{courier}
\usepackage{subfigure}
\usepackage{epstopdf}
\usepackage{algorithm}
\usepackage{algorithmic}
\usepackage{epstopdf}
\usepackage{amsthm}
\usepackage{multicol}
\usepackage{multirow}
\usepackage{braket}
\usepackage{array}
\usepackage{tabu}
\usepackage{amsmath,amssymb,amsfonts}
\usepackage{textcomp}
\usepackage{xcolor}
\usepackage{nameref}
\usepackage{amssymb}
\usepackage{makecell}
\usepackage{diagbox}
\usepackage{color}
\usepackage[normalem]{ulem}

\newcommand{\etal}{\textit{et al.}}

\newcommand{\equalcontribution}{\textsuperscript{*}}
\newcommand{\correspondingauthor}{\textsuperscript{$\dagger$}}
\renewcommand{\footnoterule}{%
  \kern -3pt
  \hrule width 0.49\textwidth height 0.4pt
  \kern 2.6pt
}

\def\BibTeX{{\rm B\kern-.05em{\sc i\kern-.025em b}\kern-.08em
    T\kern-.1667em\lower.7ex\hbox{E}\kern-.125emX}}
\begin{document}

\title{Robust Deep Learning Models against \\ Semantic-Preserving Adversarial Attack
}

\author{\IEEEauthorblockN{1\textsuperscript{st} Dashan Gao\equalcontribution\correspondingauthor
\protect\thanks{*These authors contributed equally to this work.}
\protect\thanks{\textsuperscript{$\dagger$}Corresponding author.}
\protect\thanks{This paper is accepted by the 2023 International Joint Conference on Neural Networks (IJCNN 2023).}
}
\IEEEauthorblockA{\textit{Dept. of CSE} \\
\textit{HKUST \& SUSTech}\\
Hong Kong, China \\
dgaoaa@cse.ust.hk}
\and
\IEEEauthorblockN{2\textsuperscript{nd} Yunce Zhao\equalcontribution}
\IEEEauthorblockA{\textit{Dept. of CSE} \\
\textit{University of Technology, Sydney \& SUSTech}\\
Shenzhen, China \\
yunce.zhao2023@gmail.com}
\and
\IEEEauthorblockN{3\textsuperscript{rd} Yinghua Yao}
\IEEEauthorblockA{\textit{Dept. of CSE} \\
\textit{University of Technology, Sydney \& SUSTech}\\
Shenzhen, China \\
yaoyinghua78@gmail.com}
\and
\IEEEauthorblockN{4\textsuperscript{th} Zeqi Zhang}
\IEEEauthorblockA{\textit{Huawei Technologies Co., Ltd.}\\
Shenzhen, China \\
xinlingqingkong@126.com}
\and
\IEEEauthorblockN{5\textsuperscript{th} Bifei Mao}
\IEEEauthorblockA{\textit{Huawei Technologies Co., Ltd.}\\
Shenzhen, China \\
maobifei@huawei.com}
\and
\IEEEauthorblockN{6\textsuperscript{th} Xin Yao}
\IEEEauthorblockA{\textit{Dept. of CSE} \\
\textit{SUSTech}\\
Shenzhen, China \\
xiny@sustech.edu.cn}
}

\maketitle

\begin{abstract}
Deep learning models can be fooled by small $l_p$-norm adversarial perturbations and natural perturbations in terms of attributes. 
Although the robustness against each perturbation has been explored, it remains a challenge to address the robustness against joint perturbations effectively.
In this paper, we study the robustness of deep learning models against joint perturbations by proposing a novel attack mechanism named Semantic-Preserving Adversarial (SPA) attack, which can then be used to enhance adversarial training.
Specifically, we introduce an attribute manipulator to generate natural and human-comprehensible perturbations and a noise generator to generate diverse adversarial noises. 
Based on such combined noises, we optimize both the attribute value and the diversity variable to generate jointly-perturbed samples.  
For robust training, we adversarially train the deep learning model against the generated joint perturbations.
Empirical results on four benchmarks show that the SPA attack causes a larger performance decline with small $l_{\infty}$ norm-ball constraints compared to existing approaches. 
Furthermore,  our SPA-enhanced training outperforms existing defense methods against such joint perturbations.
\end{abstract}


\section{Introduction}

\begin{figure}[t!]
  \centering
  \includegraphics[width=\columnwidth]{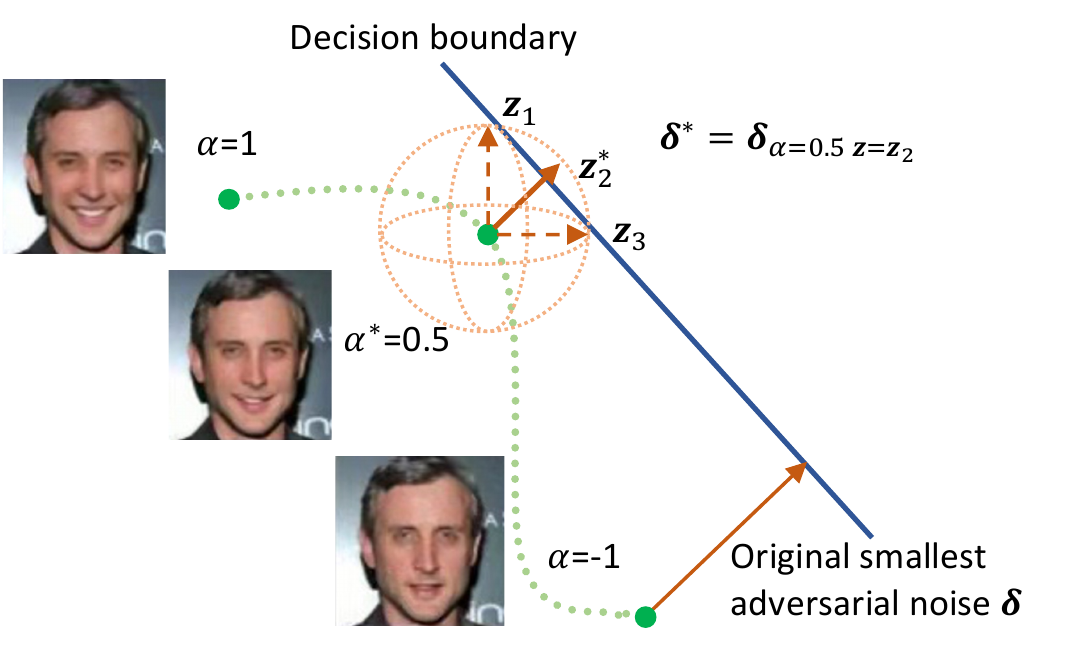}
  \caption{An illustration of SPA attack. The attribute value $\alpha$ (e.g., the extent of smiling) and adversarial diversity variable $\bm{z}$ are jointly optimized in attribute space and adversarial space to find a valid jointly-perturbed sample with smaller $l_p$-norm adversarial noises $\bm{\delta}$.}
  \label{idea_spa}
\end{figure}

Deep neural networks (DNNs) have achieved significant breakthroughs in a variety of domains and tasks, such as computer vision \cite{he2016deep,krizhevsky2012imagenet}, and natural language processing \cite{sutskever2014sequence}. 
However, DNNs are found to be susceptible to adversarial samples, which are crafted by adding small adversarial perturbations to natural samples \cite{goodfellow2014explaining}. 
Later studies suggested that such vulnerability of DNNs exists over a wide range of security-sensitive applications \cite{2016arXiv160702533K}, which highlights the significance of building robust deep learning models.

Robust deep learning aims to make accurate predictions on unseen samples with perturbations. 
Adversarial training \cite{madry2018towards, goodfellow2014explaining} proves to be a most effective strategy against adversarial samples \cite{athalye2018obfuscated, pang2021bag, NEURIPS2020_11f38f8e} and is widely adopted for building robust deep learning models \cite{kurakin2016adversarial, zhang2019theoretically, wang2020improving}.
The idea of adversarial training is to train the target model by incorporating adversarial samples. 
The adversarial training process mimics adversaries to constantly attack the target model. And the target model gradually learns to be robust against all attacks (i.e., adversarial samples) from the adversary. 
Therefore, it is essential to find threatening adversarial samples with subtle adversarial noise, which could effectively cause mis-predictions. 
\textit{This work enhances the robustness of deep learning models against different perturbations by proposing a novel and strong attack mechanism for adversarial training.}

Although most existing works on adversarial attacks aim to find imperceptible $l_p$-norm adversarial noises, adversarial noises added to data are large to perceptible in many cases (e.g., $\epsilon=0.3$ for MNIST~\cite{726791} dataset). 
Otherwise, the adversarial attack may fail. 
Therefore, existing works usually consider a relatively large size of $l_p$ ball to achieve successful adversarial attacks. 
This work designs an attack mechanism to find smaller yet viable adversarial perturbations.

Besides malicious adversarial perturbations, DNNs might also encounter \textit{natural perturbations} measured by attributes in the real world.
For example, object-level transforms, such as the degree to which a person smiles, or geometric transforms, such as the rotation of images, that commonly appear in the real world are not accounted for by adversarial perturbations. 
Natural perturbations along these attributes preserve the semantic classification information and are thus \textit{semantic-preserving}. 
For instance, changing the extent of smiling of a person in a gender classification task, or changing the rotation angle of a digit in a digit classification task, generates a natural-looking image and will not lead to a change of the true class label. 
Yet, authors in \cite{joshi2019semantic,gokhale2021attribute} find that intentional perturbations along these attributes are also likely to cause model performance to decline.
Although such performance decline is much smaller compared to that of adversarial perturbations, natural perturbations can be used for attacks.
Moreover, it is shown that the robustness against natural perturbations is independent of adversarial robustness \cite{Laugros_2019_ICCV}.
That is, classifiers trained with only adversarial samples are not robust against natural perturbations, and vice versa. Therefore, we aim to train a target model that is robust against both types of perturbations.

In this work, we address the robustness against both perturbations by proposing a novel generator-based adversary named Semantic-Preserving Adversarial (SPA) attack that generates \textit{jointly-perturbed} samples. 
Figure \ref{idea_spa} illustrates the idea of the proposed attack. 
SPA attack maximizes the exposure of the target model to variations in both attribute space and adversarial space.
The attack framework consists of an attribute manipulator for natural perturbations and an adversarial noise generator for diverse adversarial perturbations. 
By modifying the class-irrelevant attributes of the images, SPA attack searches semantic-preserving samples that are more vulnerable to adversarial attacks.
Then, SPA attack finds valid adversarial noises under stringent $l_p$ norm-ball constraints by exploring the adversarial diversity variable.
Based on the proposed attack mechanism, we further design a robust training approach, which addresses a min-max optimization and adversarially trains the target model against joint perturbations. 
Empirical studies in Section~\ref{experiments} verify the effectiveness of our SPA attack, and demonstrate that our SPA training can provide superior protection against joint attacks compared to previous methods.

The major contributions are summarized as follows:
\begin{itemize}
    \item We present a novel attack mechanism named Semantic-Preserving Adversarial (SPA) attack that considers the problem of robustness against joint perturbations in the pixel space as well as a set of specified attributes in the attribute space. 
    \item We propose SPA training as robust training by solving a min-max optimization problem and jointly exploring the pixel space and the attribute space in novel ways without access to the test domain. 
    \item We introduce two surrogate functions for attribute manipulation on two classical semantic-preserving natural perturbations: geometric transformations and objective-level transformations.
    \item We empirically demonstrate the effectiveness of our proposed approach on four public datasets. 
\end{itemize}

The rest of this paper is organized as follows. 
We briefly review the related work in Section \ref{related_works}
and provide the definition of this problem in Section \ref{background}. Then, we introduce our approach for attack and defense in Section \ref{approach}. Finally, we discuss the experiment results on MNIST, FashionMNIST, CelebA, and SICAPv2 datasets in Section \ref{experiments}.

\section{Related Work}\label{related_works}
In this section, we investigate the related work on adversarial attacks and training, attribute robust training, and attribute manipulation.

\subsection{Adversarial Attacks and Training}
As a pioneering work, Szegedy~\etal~\cite{szegedy2013intriguing} initially observed that deep learning models are vulnerable against imperceptible \textit{adversarial perturbations}. While adversarial training could be formulated as a min-max optimization mathematically, it is a huge challenge to solve the inner maximization~(e.g., the generation of adversarial examples).  Goodfellow~\etal~\cite{goodfellow2014explaining} proposed Fast Gradient Sign Method (FGSM) to generate adversarial examples with the sign of gradient in a one-step manner. However, the model trained with FGSM-generated adversarial examples suffers from catastrophic overfitting ~\cite{Wong2020Fast,kim2021understanding}.
For further enhancing the strength of attacks, an iterative FGSM variant Basic Iterative Method~(BIM), was proposed in~\cite{2016arXiv160702533K}. And CW attack~\cite{Carlini2017CandW} then took a direct optimization approach to find adversarial samples, which broke the distillation knowledge defense for the first time. 
Later, Madry~\etal~\cite{madry2018towards} proposed PGD attack, which is considered to be one of the most powerful first-order attacks for approximating the optimal value of the inner maximization problem. 

More recently, approaches based on learning-to-learn (L2L) \cite{jiang2021learning,wang2019direct,mopuri2018nag} are proposed to improve adversarial training. 
L2LDA~\cite{jang2019adversarial} further 
introduced a diversity variable to generate diverse adversarial noises
and  update the adversarial noise recursively. 
Their adversarially trained ResNet was shown to outperform the ResNets trained by CW, L2L and PGD on CIFAR-10. 
Such adversarial attacks perturb the pixel space under $l_p$ norm-ball constraints, yet may fail when the constraints get much more stringent. 
On the contrary, we approach the problem of generating jointly-perturbed samples with smaller adversarial perturbations. 
Following the literature, we use several state-of-the-art attacks from CW~\cite{Carlini2017CandW} and L2LDA~\cite{jang2019adversarial} as our benchmarks. 

\subsection{Attribute Robust Training}
Recently, there has been an increasing interest in the robustness against \textit{natural perturbations} in terms of attributes, which are perceptible shifts in the data but still are natural-looking, and enough to fool a classifier ~\cite{hendrycks2018benchmarking}. 
Liu~\etal~\cite{liu2018beyond} proposed to perturb physical parameters that underlay image formation to produce natural perturbations sensitive to physical concepts like lighting and geometry.
\cite{joshi2019semantic} proposed to generate natural perturbations by modifying multiple specified attributes with a conditional generative model.
\cite{gokhale2021attribute} proposed to perturb the attribute space to synthesize new images and maximize the exposure of the classifier to the attributes space.
However, such attacks only cause \textit{limited} performance decline and do not result in severe model failure as pixel-level adversarial attacks do~\cite{gokhale2021attribute}, especially when few attributes are valid. 
Contrary to these approaches, we address the robustness against both perturbations by considering a novel and powerful attack that jointly leverages attribute perturbations and adversarial perturbations. 

\begin{figure*}[t!]
  \centering
  \includegraphics[width=\textwidth]{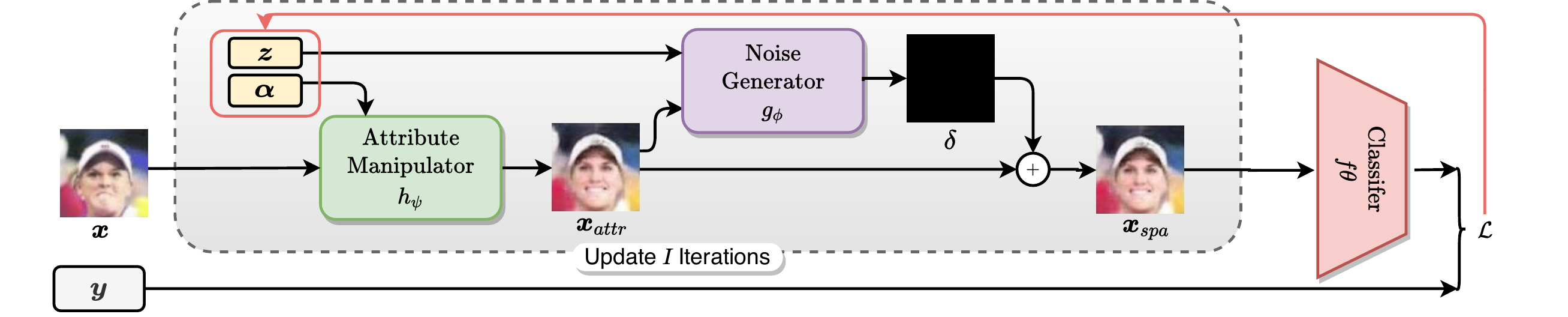}
  \caption{An illustration of the Semantic-Preserving Adversarial (SPA) attack. The attribute manipulator $h_{\psi}$ and the noise generator $g_{\phi}$ sequentially conduct natural perturbations and adversarial perturbations. The attribute value $\bm{\alpha}$ (e.g., the extent of smiling) and adversarial diversity variable $\bm{z}$ are jointly and iteratively optimized to generate jointly-perturbed samples.}
  \label{pipeline_spa}
\end{figure*}

\subsection{Attribute Manipulation}
There are various approaches for manipulating attributes in images. Conditional Generative Adversarial Networks \cite{mirza2014conditional} use an encoder-decoder architecture to learn attribute invariant latent representation and disentangle attributes for attribute editing. Matrix Subspace Projection (MSP)~\cite{li2020latent} is proposed to factorize the latent space for attribute manipulation effectively. Spatial Transformer Networks (STN)~\cite{jaderberg2015spatial} uses a localization net to conduct precise spatial transformation.

\section{Problem Formulation}\label{background}

We begin by defining a classifier parameterized by $\theta$ as $f_{\theta}: \mathcal{X} \rightarrow \mathcal{Y}$, where $\mathcal{X}$ denotes the space of the image data and $\mathcal{Y}$ denotes the label space. 
Labeled data $(\bm{x}, \bm{\alpha}, \bm{y})$, where $\bm{x}$ denotes the input image, $\bm{\alpha}$ denotes the annotated attribute value, $\bm{y}$ denotes the class label. Table~\ref{table_notions} summarizes the notions used in this paper.

The objective of this paper is to build a classifier $f_{\theta}$ that is robust to both 1) \textit{adversarial} perturbations, which are $l_{\infty}$-bounded in the pixel space, and 2) \textit{natural} perturbations along specific attributes $\bm{\alpha}$ that is specified a \textit{priori}. 

We categorize semantic-preserving natural perturbations 
into two major types - (a) \textbf{Geometric transformations}, where images are manipulated by affine transformations such as rotation, scaling, and shifting. 
(b) \textbf{Object-level transformations}, where general attributes of the objects in images are manipulated without changing the semantic information of the class labels, such as changing the extent of smiling or hair color of a person in a gender classification task. 
Geometric transformations have explicit generative mechanisms, allowing us to conduct accurate natural perturbations. 
However, we do not have access to the true generative mechanisms for objective-level transformations. Therefore, we need to approximate the objective-level transformations by training a conditional generative model. 

The classification loss of the target classifier $f_{\theta}$ can be defined as a cross-entropy loss:
\begin{equation}
    \mathcal{L} (\bm{x}, \bm{y}; \theta) = \bm{y}^T \log f(\bm{x}; \theta),
\end{equation}
where $f(\bm{x}; \theta)$ outputs the predicted probabilities for $\bm{x}$. We simplify the loss as $\mathcal{L} (\bm{x})$ without causing confusion.

This task can be formulated as a min-max optimization of a saddle point problem:
\begin{equation}\label{eq_minmax1}
    \min_{\theta \in \Theta} \mathbb{E}_{(\bm{x},\bm{y}) \sim \mathcal{D}} \left[ \max_{\tilde{\bm{x}} \in \mathcal{X}_{spa}(\bm{x})} \mathcal{L} (\tilde{\bm{x}}, \bm{y}; \theta) \right],
\end{equation}
where 
$\mathcal{X}_{spa}(\bm{x})$ is a set of admissible attacks of $\bm{x}$ with natural perturbations in the attribute space along the specified attributes $\bm{\alpha}$ as well as $l_{p}$-bounded adversarial perturbations in the pixel space. 

\begin{table}[t]
\centering
\caption{Notations}
\begin{tabular}{ c | l } 
 \toprule
 Notations & Descriptions \\
 \midrule
 \midrule
 $\bm{x}$ & The input features \\
 $\bm{y}$ & The labels \\
 $\bm{\alpha}$ & The attribute values \\
 $\bm{z}$ & The diversity variable of adversarial noises \\
 $f_{\theta}$ & The classifier parameterized by $\theta$ \\
 $g_{\phi}$ & The adversarial noise generator parameterized by $\phi$ \\
 $h_{\psi}$ & The attribute manipulator parameterized by $\psi$ \\
 $\mathcal{L}$ & The loss function \\
 $\epsilon$ & The $l_{\infty}$ norm of adversarial noise \\
 \bottomrule
\end{tabular}

\label{table_notions}
\end{table}

Eq.~\ref{eq_minmax1} solves a min-max optimization problem. The inner maximization aims to generate jointly-perturbed samples with both perturbations that maximize the classification loss, and the outer minimization corresponds to finding model parameters that minimize the loss on jointly-perturbed samples found by the inner maximization. 
The success of finding the optimal $\theta^{*}$ that minimizes Eq.~\ref{eq_minmax1} crucially relies on solving the  (non-concave) inner optimization problem. 
In this work, we focus on the inner maximization problem of Eq.~\ref{eq_minmax1} to achieve robustness against both perturbations. To conduct attacks, we propose an architecture that generates and effectively optimizes diverse natural and adversarial perturbations.

\section{Proposed Approach}\label{approach}
Conceptually, generating jointly-perturbed samples with both natural and adversarial perturbations of an image can be broken into two sub-problems: (a) effective navigation on the attribute space and the $l_p$-bounded pixel space, and (b) joint optimization of attribute values and adversarial noises over both attribute and pixel spaces. We address each problem in detail below. 

\subsection{Joint Perturbation Models}
First, let us consider the problem of generating \textit{jointly-perturbed} samples with both natural and adversarial perturbations, via navigating on the attribute space and the $l_p$-bounded pixel space. 
We design a sequential architecture to perform joint perturbation, as shown in Figure~\ref{pipeline_spa}. 

\textbf{Attribute Manipulation}. We adopt a differentiable attribute manipulator $h_{\psi}: \mathcal{X} \times \mathcal{A} \rightarrow \mathcal{X}$, conditioned on a semantic attribute space $\mathcal{A}$, to conduct attribute manipulation. 
$h_{\psi}$ takes an image $\bm{x}$ and an attribute vector $\bm{\alpha}$ as input, and generates perturbed image $\bm{x_{attr}}$ with the specified attribute. $h_{\psi}$ is trained with the annotated training data $(\bm{x}, \bm{\alpha})$. 

For objective-level transformations, we leverage MSP~\cite{li2020latent} to build $h_{\psi}$. An encoder network encodes the input image into attribute-invariant features and attribute-related features. Matrix subspace projection modifies the attributes of interest to specified values $\bm{\alpha}$. Then, a decoder network decodes the modified features into the desired image with specified attributes. 
For geometric transformations, the attribute manipulator is built with STN~\cite{jaderberg2015spatial}. 
The input images are first normalized by STN with attributes predicted by an attribute predictor, which is pre-trained with the annotated attributes in training data.
Then an STN is used to conduct affine transformation on the normalized images to specified attributes. 

\textbf{Adversarial Noise Generation}. 
To explore novel and hard adversarial samples, we explicitly train an adversarial noise generator $g_{\phi}: \maketitle{X} \times \mathcal{Z} \rightarrow \mathcal{S}$ to generate diverse adversarial noises $\delta \in \mathcal{S}$, conditioned on a diversity variable $\bm{z} \in \mathcal{Z}$, where $\mathcal{S}$ the $l_p$-ball with $\epsilon$ size and $\mathcal{Z}$ the diversity variable space. 
For a stronger attack, the adversarial noise generator $g_{\phi}$ can be further enhanced by adding $\nabla_{\bm{x}}\mathcal{L}(\bm{x})$ the gradient of the image, $\bm{y}$ the class label, and $\delta$ the noise of the image as input.
Therefore, the adversarial noise can be recursively generated by:
\begin{equation}\label{eq_adv_noise_gen}
    \tilde{\bm{x}}^{(t+1)} \leftarrow \text{Proj}_{\mathcal{S}\cup \mathcal{X}}(\bm{x}^{(t)} + \epsilon_{\text{step}} g_{\phi}(\bm{x}, \bm{z}; \bm{y}, \delta^{(t)}, \nabla_{\bm{x}}\mathcal{L}(\tilde{\bm{x}}^{(t)})) ),
\end{equation}
where $\text{Proj}_{\mathcal{S}\cup \mathcal{X}}(\cdot)$ denotes the projection of its element to $l_{p}$-ball $\mathcal{S}$ and a valid pixel value range,  $\delta^{(t)}$ is the noise accumulated up to $t$-th step, and $\epsilon_{step}$ denotes the step size smaller than $\epsilon$. 
We adopt a diversity loss to encourage generating diverse adversarial noises~\cite{yang2019diversity,jang2019adversarial}:
\begin{equation}\label{eq_div_loss}
    \mathcal{L}_{div} = \frac{\frac{1}{T}\sum^{T}_{1} ||\tilde{\bm{x}}^{(t)}(\bm{z}_1) - \tilde{\bm{x}}^{(t)}(\bm{z}_2) ||_{1}}{|| \bm{z}_1 - \bm{z}_2 ||_1},
\end{equation}
where $\tilde{\bm{x}}^{(t)}(\bm{z})$ denotes the adversarial samples generated by Eq.~\ref{eq_adv_noise_gen} with $\bm{z}$, and $\bm{z}_1, \bm{z}_2$ are two i.i.d. samples of $\bm{z}$.

Given the above models, we define a semantic-preserving adversarial attack as the process of transforming an input image $x$ via attribute perturbation $\bm{x}_{\bm{\alpha}} = h_{\psi}(\bm{x}, \bm{\alpha})$ and adversarial perturbation $\delta = g_{\phi}(\bm{x}_{\bm{\alpha}}, \bm{z})$ to produce a new sample $\widetilde{\bm{x}_{\bm{\alpha}}} = \bm{x}_{\bm{\alpha}} + \delta$, such that $f(\widetilde{\bm{x}_{\bm{\alpha}}}) \neq y$. 
Therefore, we reformulate Eq.~\ref{eq_minmax1} as an optimization of the following problem: 
\begin{align}\label{eq_minmax}
    &\min_{\theta \in \Theta} \mathbb{E}_{(\bm{x},y) \sim \mathcal{D}} \left[ \max_{\bm{\alpha} \in \mathcal{A}, z \in \mathcal{Z}} \mathcal{L} (\bm{x}_{spa}, y; \theta) \right] \\
    & s.t. \;\; \bm{x}_{spa} = \text{Proj}_{\mathcal{S}\cup \mathcal{X}} (h_{\psi}(\bm{x}, \bm{\alpha}) + \epsilon g_{\phi}(h_{\psi}(\bm{x}, \bm{\alpha}), \bm{z})), \notag
\end{align}

\begin{algorithm} [t!]
\caption{Semantic-Preserving Adversarial (SPA) Attack}
\begin{algorithmic}[1]\label{algo_spa_attack}
\REQUIRE Image $\bm{x}$, ground-truth label $\bm{y}$, pretrained attribute manipulator and noise generator $h_{\psi}, g_{\phi}$.
\ENSURE SPA sample $\bm{x}_{spa}$.
\STATE Initialize $\bm{\alpha}^{(0)}$, $\bm{z}^{(0)}$
\FOR{i in range [0, I)}
\STATE $\bm{x}_{attr}^{(i)} \leftarrow h_{\psi}(\bm{x}, \bm{\alpha}^{(i)})$. 
\FOR{t in range [0, T]}
\STATE $\tilde{\bm{x}}_{attr}^{(i, t+1)} \leftarrow \text{Proj}_{\mathcal{S}\cup \mathcal{X}}(\bm{x}_{attr}^{(i)} + $ \\
$\;\;\;\;\;\;\;\;\; \epsilon_{\text{step}} g_{\phi}(\bm{x}_{attr}^{(i)}, \bm{z}^{(i)}; \bm{y}, \delta^{(t)}, \nabla_{\bm{x}}\mathcal{L}(\tilde{\bm{x}}_{attr}^{(i, t)})) )$
\ENDFOR 
\STATE $\tilde{\bm{x}}_{attr}^{(i+1)} \leftarrow \tilde{\bm{x}}^{(i, T)}$
\STATE $\bm{\alpha}^{(i+1)} \leftarrow \bm{\alpha}^{(i)} + \nabla_{\bm{\alpha}}\mathcal{L}(\tilde{\bm{x}}_{attr}^{(i+1)})$
\STATE $\bm{z}^{(i+1)} \leftarrow \bm{z}^{(i)} + \nabla_{\bm{z}}\mathcal{L}(\tilde{\bm{x}}_{attr}^{(i+1)})$
\ENDFOR
\STATE $\bm{x}_{spa} \leftarrow $ Repeat Step 3-6 with $\bm{\alpha}=\bm{\alpha}^{(I)}, \bm{z}=\bm{z}^{(I)}$.
\end{algorithmic}
\end{algorithm}

\subsection{Iterative Parameter Optimization}
Having access to the attribute manipulator and the noise generator, we focus on solving the inner optimization in Eq.~\ref{eq_minmax}. 
Note that the success of robust training relies on generating strong perturbations in terms of attributes and adversarial noises. 
Algorithm~\ref{algo_spa_attack} demonstrates the optimization of the attribute value and diversity variable for finding SPA samples $\bm{x}_{spa}$.
Step 3-6 conducts attribute manipulation and adversarial noise generation sequentially. 
Then, we project the adversarial loss onto the attribute space and the diversity variable space, by cascading the output of the attribute manipulator, noise generator, and target classifier. 
The optimization is conducted by back-propagation over the classifier $f_{\theta}$ and perturbation models $h_{\psi}, g_{\phi}$. 

Algorithm~\ref{algo_spa_attack} efficiently explores a larger parameter space and generates stronger jointly-perturbed samples, compared to previous methods~\cite{jang2019adversarial,gokhale2021attribute}, especially under stringent constraints with small $l_p$-ball and few valid semantic-preserving attributes. 

\subsection{Semantic-Preserving Adversarial Training}
With our proposed SPA attack, we solve the outer optimization problem in Equation~\ref{eq_minmax} by proposing a unified training framework for the robustness against natural and adversarial perturbations, by jointly optimizing the noise generator $g_{\phi}$ and the target classifier $f_{\theta}$. Before SPA training, The attribute manipulator $h_{\psi}$ is pretrained with the annotated training data $(\bm{x}, \bm{\alpha}_{gt})$, where $\bm{\alpha}_{gt}$ the ground-truth attributes. 

\begin{algorithm} [h]
\caption{Semantic-Preserving Adversarial Training}
\begin{algorithmic}[1]\label{algo_spa_training}
\REQUIRE Image $\bm{x}$, and label $\bm{y}$, pretrained attribute manipulator $h_{\psi}$.
\STATE $\bm{x}_{spa} \leftarrow \text{SPA\_Attack}(\bm{x}, \bm{y}, h_{\psi}, g_{\phi})$
\STATE Randomly initialize $\bm{\alpha}, \bm{z}_1, \bm{z}_2$
\STATE $\bm{x}\leftarrow [\bm{x}; \bm{x}], \bm{\alpha} \leftarrow [\bm{\alpha}; \bm{\alpha}], \bm{y}\leftarrow [\bm{y}; \bm{y}], \bm{z}\leftarrow[ \bm{z}_{1}; \bm{z}_{2}] $ ($[\cdot ; \cdot]:$ concatenation)
\STATE $\bm{x}_{attr} \leftarrow h_{\psi}(\bm{x}, \bm{\alpha})$
\FOR{t in range [0, T)}
\STATE $\tilde{\bm{x}}_{attr}^{(t+1)} \leftarrow \text{Proj}_{\mathcal{S} \cup \mathcal{X}}(\bm{x}_{attr} + $ \\
$\;\;\;\;\;\;\;\;\; \epsilon_{\text{step}} g_{\phi}(\bm{x}_{attr}, \bm{z}; \bm{y}, \delta^{(t)}, \nabla_{\bm{x}}\mathcal{L}(\tilde{\bm{x}}_{attr}^{(t)})) )$
\ENDFOR
\STATE Compute $\mathcal{L}_{div}$ with $\tilde{\bm{x}}_{attr}^{(t)}, \bm{z}$ following Eq.~\ref{eq_div_loss}
\STATE $\mathcal{L}_{cls} \leftarrow \frac{1}{T} \sum^{T}_{1} \mathcal{L}(\tilde{\bm{x}}_{attr}^{(t)}) + \mathcal{L}(\bm{x}_{spa})$
\STATE $\phi \leftarrow \phi + \nabla_{\phi} ( \mathcal{L}_{cls} + \lambda \mathcal{L}_{div} )$
\STATE $\theta \leftarrow \theta - \nabla_{\theta}(\mathcal{L}(\bm{x}) + \mathcal{L}(\tilde{\bm{x}}_{attr}^{(T)}) + \mathcal{L}(\bm{x}_{spa}))$
\end{algorithmic}
\end{algorithm}

The overall procedure is summarized in Algorithm~\ref{algo_spa_training}. 
For each batch, 
we first randomly sample attribute values for attribute manipulation and two batches of diversity variables for diverse adversarial noises.
Then, we duplicate the batch with different diversity variables.
Then, we conduct attribute manipulation on the training data and generate diverse adversarial noise with different diversity variables.
Then, we generate SPA samples of training data following Algorithm~\ref{algo_spa_attack}.
We update the noise generator $g_{\phi}$ with the classification loss $\mathcal{L}_{cls}$ and the diversity loss $\mathcal{L}_{div}$. 
Finally, we update the target classifier $f_{\theta}$ with the classification loss of real images, attribute perturbed images, and SPA images.

\section{Experiments}\label{experiments}
In this section, we report the experimental results of our proposed SPA attack and training on four public datasets.

\subsection{Experimental Setups}
\textbf{Datasets.} We evaluate our approach on MNIST~\cite{726791}, FashionMNIST~\cite{xiao2017fashion}, CelebFaces Attributes (CelebA)~\cite{liu2015deep} and SICAPv2~\cite{Silva_Rodr_guez_2020} datasets. 
1) \textbf{MNIST} dataset consists of $70,000$ $28 \times 28$ images with digits from zero to nine. 
2) \textbf{Fashion-MNIST} dataset consists of $70,000$ $28 \times 28$ images with 10 classes of clothes. 
For both datasets, we consider rotation in $[-45, 45]$ degrees and scaling in $[0.7, 1.3]$ times as natural perturbations, separately.

We also conduct experiments on two real-world datasets.
3) \textbf{CelebA} dataset has more than 200K $64\times64$ celebrity images, each with 40 attribute annotations.
We consider the "smiling" attribute as the target attribute for natural perturbation and train a classifier to predict "gender". 
All training data are resized to $32\times32$. 
4) \textbf{SICAPv2} is a medical dataset collected for prostate cancer diagnosis. There are 3773 non-cancerous patches and 3641 cancerous patches with Gleason grading equal 4. We resize the images to $64 \times 64$ and consider the rotation in $[-45, 45]$ degrees as natural perturbations. 
We set the size of $l_{\infty}$ ball as $\epsilon=0.1$ for MNIST and FashionMNIST, and $\epsilon=0.01$ for CelebA and SICAPv2 datasets, which is $\mathbf{3\times}$ \textit{smaller} than that of previous works. 

\textbf{Model Parameters.}
We employ ResNet-20 as the target classifier. 
We randomly initialize the classifier and the generator and jointly train both of them by Adam with a learning rate of 0.01 beta of $[0.9, 0.99]$ and weight decay of 0.0001 for 20K iterations, with a batch size of 100. 
For SPA attack, after the classifier is pre-trained, the generator is updated for 100K iterations, with a batch size of 100. 
We set $\epsilon_{step}=1/4 \epsilon$.
We update the attribute and adversarial noise to generate SPA samples for ten steps. 
Both attribute value and diversity variable are optimized by Adam over the cross-entropy loss. 

\textbf{Baselines}. 
To measure the robustness of the SPA-trained classifier, we compare our approach against the classifiers that are trained upon state-of-the-art adversarial training methods. We employ two gradient-based methods CW~\cite{Carlini2017CandW} and PGD~\cite{madry2018towards}, and a generator-based method L2LDA\cite{jang2019adversarial} for adversarial training. We also compare our approach against an attribute robust training method AGAT~\cite{gokhale2021attribute}.

To evaluate the effectiveness of the SPA attack, we compare our SPA attack with a natural perturbation method SAA~\cite{joshi2019semantic}, and three adversarial perturbation approaches CW, PGD, and L2LDA. 

Furthermore, to demonstrate the effectiveness of the proposed joint optimization of attribute values and diversity variables, we also compare with hybrid approaches which combine existing adversarial attack approaches, including CW, PGD and L2LDA with random attribute perturbation, denoted as CW-Attr, PGD-Attr and L2LDA-Attr, correspondingly. 

\begin{table}[ht]
\centering
\caption{Impact of the perturbation models on SPA attack on FashionMNIST dataset with "rotation" attribute, in terms of accuracy. 
$\times$ denotes removing the model $h_{\psi}$ or $g_{\phi}$.}
\begin{tabular}{ c | c c c c } 
 \toprule
 \diagbox{Attack}{Defense} & Naive & AGAT & CW & SPA \\
 \midrule
 \midrule
 SPA ($h_{\psi} \times \; g_{\phi} \times$) & 0.9337 & 0.9153 & 0.9241 & 0.9251 \\
 SPA ($h_{\psi}\; \checkmark \; g_{\phi} \times$) & 0.7451 & 0.8541 & 0.7544 & 0.8849 \\
 SPA ($h_{\psi} \times \; g_{\phi} \checkmark$) & 0.0737 & 0.1298 & 0.5190 & 0.8428 \\
 SPA ($h_{\psi} \;\checkmark \; g_{\phi}\;\checkmark$) & \textbf{0.0093} & \textbf{0.0189} & \textbf{0.5092} & \textbf{0.7996} \\
 \bottomrule
\end{tabular}

\label{ablation_model}
\end{table}

\begin{table}[h]
\centering
\caption{Impact of the parameter optimization on SPA attack on FashionMNIST dataset with "rotation" attribute, in terms of accuracy. $\bm{\alpha}$ and $\bm{z}$ refer to the attribute value and the diversity variable in Algorithm~\ref{algo_spa_attack}. $\times$ denotes 
using fixed random $\bm{\alpha}$ or $\bm{z}$.}
\begin{tabular}{ c | c c c c } 
 \toprule
 \diagbox{Attack}{Defense} & Naive & AGAT & CW & SPA \\
 \midrule
 \midrule
 SPA ($\bm{\alpha} \times \; \bm{z}\;\times$) & 0.0541 & 0.0863 & 0.5144 & 0.8327 \\
 SPA ($\bm{\alpha}\; \checkmark \; \bm{z}\;\times$) & 0.0144 & 0.0311 & 0.5102 & 0.8209 \\
 SPA ($\bm{\alpha} \times \; \bm{z}\;\checkmark$) & 0.0284 & 0.1045 & \textbf{0.5084} & 0.8255 \\
 SPA ($\bm{\alpha}\; \checkmark \; \bm{z}\;\checkmark$) & \textbf{0.0093} & \textbf{0.0189} & 0.5092 & \textbf{0.7996} \\
 \bottomrule
\end{tabular}

\label{ablation_parameter}
\end{table}

\begin{figure}[t!]
  \centering
  \includegraphics[width=0.75\columnwidth]{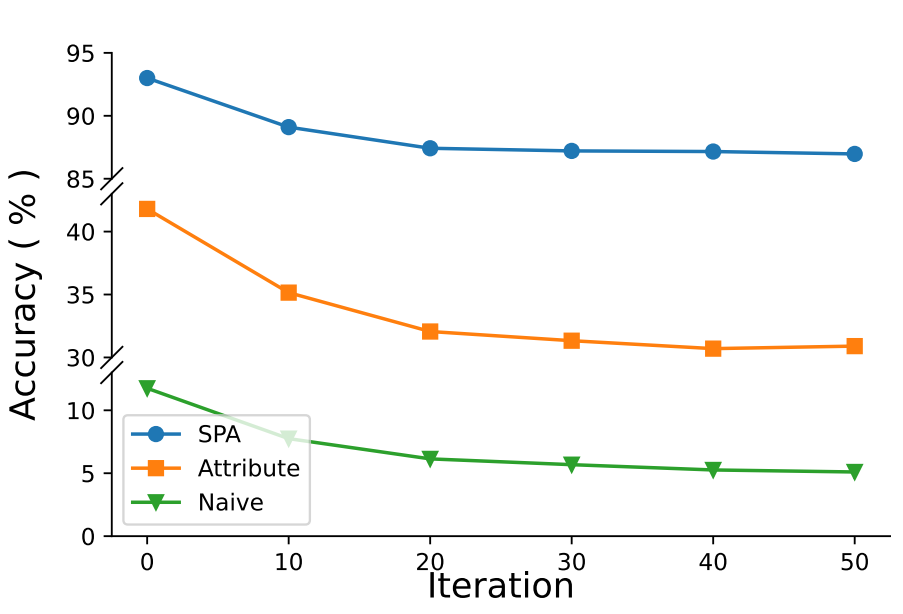}
  \caption{Impact of attack iterations on classification accuracy on CelebA dataset with "Smiling" attribute. We perform SPA attack method on naive,
  AGAT~\cite{gokhale2021attribute}, and SPA-trained classifiers.}
  \label{spa_update}
\end{figure}

\subsection{Ablation Study}
To understand (a) how the perturbation models impact the performance of SPA attack, and (b) how the optimization of the attribute value $\bm{\alpha}$ and the diversity variable $\bm{z}$ improves the quality of SPA samples, we conduct ablation studies on the FashionMNIST dataset, using STN-based attribute manipulator. Note that STN conducts precise natural perturbation along with attributes and thus will not unintentionally introduce an adversarial noise. Therefore, the noise $\bm{\delta}$ generated by the noise generator is the overall adversarial noise added to the data. 
This allows us to \textit{isolate} the influence of each perturbation in SPA attacks.

\textbf{Perturbation Model.}
Table~\ref{ablation_model} demonstrates the impact of each perturbation model on SPA attack. We compare the second row and the third row with the first row, respectively. We observe that the adversarial noise generator $g_{\phi}$ causes severe model failure (85\% accuracy decline against naive classifier on FashionMNIST), while the attribute manipulator $h_{\psi}$ only results in 18.86\% accuracy decline.
Therefore, the adversarial noise generator contributes more accuracy decline than the attribute manipulator against the four defense methods. 

\begin{table}[h]
\centering
\caption{Impact of the noise size $\epsilon$ on CelebA dataset with "smiling" attribute, in terms of accuracy. We conduct PGD, L2LDA, and SPA attacks on a naive classifier and a SPA-trained classifier.}
\begin{tabular}{c | c | c c c c } 
 \toprule
  Defense & Attack & \multicolumn{4}{c}{Noise size $\epsilon$} \\
 \midrule
 \midrule
   & & 0.00 & 0.01 & 0.02 & 0.03 \\
 \midrule
 \multirow{3}{*}{Naive} & PGD   & 0.9565 & 0.0935 & 0.0225 & \textbf{0.0027} \\
 & L2LDA & 0.9565 & 0.1055 & 0.0372 & 0.0108 \\
 & SPA   & 0.9565 & \textbf{0.0755} & \textbf{0.0214} & 0.0051 \\
 \midrule
  \multirow{3}{*}{SPA} & PGD   & 0.9430 & 0.9130 & 0.7315 & 0.5792 \\
 & L2LDA & 0.9430 & 0.9185 & \textbf{0.5721} & 0.4744 \\
 & SPA   & 0.9430 & \textbf{0.8910} & 0.6122 & \textbf{0.4530} \\
 \bottomrule
\end{tabular}

\label{ablation_noisesize}
\end{table}

\begin{table}[h]
\centering
\caption{Impact of the attribute number on SICAPv2 dataset, in terms of accuracy. Legend of attributes: Re: reshape, Sc: scaling, Ro: rotation. We conduct the attack on a PGD-trained classifier. 
Rand-Attr is an adversarial attack with randomly generated attributes. 
As the number of attributes increases, SPA attacks are more effective. Our SPA attack performs better than Rand-Attr, where the attribute values are randomly selected, showing the efficacy of joint optimization in finding strong adversarial samples.}
\begin{tabular}{c | c | c c } 
 \toprule
  Attack Type & Attributes & SPA & Rand-Attr \\
 \midrule
 \midrule
 \multirow{3}{*}{Single Attribute} & Re & 0.3030 & 0.4136 \\
 & Sc & 0.5741 & 0.6277 \\
 & Ro & 0.3732 & 0.3830 \\
 \midrule
  \multirow{3}{*}{Double Attributes} & Re, Sc & 0.2944 & 0.3541 \\
 & Re, Ro & 0.2629 & 0.3842 \\
 & Sc, Ro & 0.3580 & 0.3732 \\
 \midrule
 Multi Attributes & Re, Sc, Ro & 0.2207 & 0.3317 \\
 \bottomrule
\end{tabular}

\label{multi_attribute}
\end{table}

\begin{table*}[ht!]
\centering
\caption{The result of White-box attacks on MNIST, FashionMNIST, CelebA, and SICAPv2 datasets. For adversarial noise, we set $\epsilon=0.1$ for MNIST and FashionMNIST datasets and $\epsilon=0.01$ for CelebA and SICAPv2 datasets. We measure the classification accuracy of the adversarially trained classifiers (rows) against various attack methods (columns).}
\begin{tabular}{ c | c c c c c c c c || c } 
 \toprule
 \diagbox{Defense}{Attack} & Naive & SAA & CW & PGD & L2LDA & CW-Attr & L2LDA-Attr & SPA & Min \\
 \midrule
 \midrule
  \multicolumn{10}{c}{MNSIT~\cite{726791} $\;\;\;$ Attribute $\leftarrow$ Scaling } \\
 \midrule
 Plain         & \textbf{0.9684} & 0.8236 & 0.1226 & 0.0983 & 0.1159 & 0.0351 & 0.0308 & 0.0118 & 0.0118 \\
 AGAT         & 0.9672 & \textbf{0.9584} & 0.1328 & 0.1240 & 0.1401 & 0.0741 & 0.1143 & 0.0219 & 0.0219 \\
 CW          & 0.9629 & 0.8307 & 0.8172 & 0.4131 & 0.7931 & 0.7497 & 0.6278 & 0.6415 & 0.6278 \\
 PGD           & 0.9891 & 0.8406 & 0.7521 & 0.7842 & 0.8028 & 0.7412 & 0.7749 & 0.6744 & 0.6744 \\
 L2LDA        & 0.9677 & 0.8472 & 0.8641 & 0.8140 & \textbf{0.9148} & 0.8073 & 0.7946 & 0.7441 & 0.7441 \\
 PGD-Attr     & 0.9860 & 0.9431 & 0.7581 & 0.7872 & 0.7992 & 0.7528 & 0.7617 & 0.7844 & 0.7581\\
 SPA  & 0.9661 & 0.9575 & \textbf{0.9194} & \textbf{0.8789} & 0.9071 & \textbf{0.8951} & \textbf{0.8784} & \textbf{0.8572} & \textbf{0.8572} \\
  \midrule
  \midrule
  
 \multicolumn{10}{c}{FashionMNSIT~\cite{xiao2017fashion}$\;\;\;$ Attribute $\leftarrow$ Rotation} \\
 \midrule
 Plain         & \textbf{0.9337} & 0.7496 & 0.0648 & 0.0617 & 0.0737 & 0.0165 & 0.0284 & 0.0093 & 0.0093 \\
 AGAT         & 0.9153 & 0.8659 & 0.0869 & 0.0794 & 0.1298 & 0.0831 & 0.1045 & 0.0189 & 0.0189 \\
 CW          & 0.9241 & 0.7597 & 0.6539 & 0.3642 & 0.5190 & 0.6831 & 0.5084 & 0.5092 & 0.5084 \\
 PGD          & 0.9257 & 0.7642 & 0.7365 & 0.7044 & 0.6531 & 0.6611 & 0.6417 & 0.5371 & 0.5371 \\
 L2LDA        & 0.9224 & 0.7827 & \textbf{0.8593} & 0.7281 & 0.7417 & \textbf{0.8144} & 0.7929 & 0.7731 & 0.7417 \\
 PGD-Attr     & 0.9217 & 0.8582 & 0.7912 & 0.7380 & 0.7278 & 0.7128 & 0.7417 & 0.6982 & 0.6982\\
 SPA  & 0.9251 & \textbf{0.8845} & 0.8313 & \textbf{0.8047} & \textbf{0.8428} & 0.8127 & \textbf{0.8255} & \textbf{0.7996} & \textbf{0.7996} \\
 \midrule
 \midrule
  
 \multicolumn{10}{c}{CelebA~\cite{liu2015deep} $\;\;\;$ Attribute $\leftarrow$ Smiling } \\
 \midrule
 Plain        & \textbf{0.9565} & 0.8830 & 0.1375 & 0.1035 & 0.1055 & 0.1035 & 0.0950 & 0.0775 & 0.0775 \\
 AGAT         & 0.9400 & \textbf{0.9420} & 0.3740 & 0.3577 & 0.4180 & 0.3725 & 0.3580 & 0.3415 & 0.3415 \\
 CW         & 0.9518 & 0.9151 & 0.5798 & 0.4966 & 0.5419 & 0.5541 & 0.5221 & 0.5077 & 0.5077 \\
 PGD           & 0.9479 & 0.9068 & 0.6431 & \textbf{0.9047} & 0.5541 & 0.5681 & 0.5328 & 0.5041 & 0.5041 \\
 L2LDA       & 0.9523 & 0.9094 & 0.6740 & 0.6937 & 0.7240 & 0.6192 & 0.6891 & 0.6390 & 0.6192 \\
 PGD-Attr     & 0.9401 & 0.9268 & 0.6458 & 0.8806 & 0.5618 & 0.5828 & 0.5527 & 0.5541 & 0.5527\\
 SPA & 0.9430 & 0.9405 & \textbf{0.9125} & 0.8973 & \textbf{0.9185} & \textbf{0.9035} & \textbf{0.8975} & \textbf{0.8910} & \textbf{0.8910} \\
 \midrule
 \midrule
  
 \multicolumn{10}{c}{SICAPv2~\cite{Silva_Rodr_guez_2020} $\;\;\;$ Attribute $\leftarrow$ Rotation } \\
 \midrule
 Plain        & \textbf{0.8971} & 0.8070 & 0.0014 & 0.0142 & 0.0217 & 0.0028 & 0.0121 & 0.0217 & 0.0014 \\
 AGAT         & 0.7624 & 0.8791 & 0.0430 & 0.0841 & 0.0402 & 0.0830 & 0.0402 & 0.0531 & 0.0402 \\
 CW           & 0.8732 & 0.7832 & 0.3044 & 0.2831 & 0.2629 & 0.3188 & 0.2370 & 0.2207 & 0.2207 \\
 PGD          & 0.8890 & 0.7890 & 0.6731 & 0.3347 & 0.4141 & \textbf{0.6542} & 0.3030 & 0.3732 & 0.3030 \\
 L2LDA        & 0.8877 & 0.8021 & 0.5844 & 0.5821 & 0.5741 & 0.5830 & 0.5703 & 0.4785 & 0.4785 \\
 PGD-Attr     & 0.8941 & 0.8741 & \textbf{0.6525} & 0.5506 & 0.4633 & 0.6277 & 0.4527 & 0.4071 & 0.4071\\
 SPA          & 0.8920 & \textbf{0.8837} & 0.6207 & \textbf{0.8533} & \textbf{0.7685} & 0.6183 & \textbf{0.7051} & \textbf{0.6295} & \textbf{0.6183} \\
 \bottomrule

\end{tabular}

\label{attack_defence_table}
\end{table*}

We further observe Table~\ref{ablation_parameter} to study the impact of parameter optimization by comparing random initialization with parameter optimization methods. 
By comparing the second and first row, we note that the optimization of attributes $\bm{\alpha}$ leads to a 1.18\% accuracy decline against the SPA defense on FashionMNIST. Similarly, we observe the performance drops 0.72\% by optimizing the diversity variable $\bm{z}$.
Therefore, optimizing attributes $\bm{\alpha}$ contributes more than optimizing the diversity variable $\bm{z}$. 

The above ablation studies show that our method's perturbation models and parameter optimizations benefit the generation of powerful attacks in a complementary way. 

\subsection{Impact of Parameters}

\textbf{Noise Size.}
As demonstrated in Table~\ref{ablation_noisesize}, we explore the noise size $\epsilon$ on the CelebA dataset with the "smiling" attribute. We conduct PGD, L2LDA, and SPA attacks on a naive classifier and a classifier adversarially trained via SPA. The size of adversarial noise varies from 0.0 to 0.03. It can be observed that when the size of noise is small (e.g., $\epsilon=0.01$), our proposed SPA attack outperforms PGD and L2LDA by around 2\%. 
As the noise size increases, the SPA attack can achieve a strong attack but could be outperformed by other approaches (e.g., PGD on a naive classifier). 

\textbf{Attack Iteration.}
To study the impact of the attack iteration $I$ in Algorithm~\ref{algo_spa_attack}, we observe the change of prediction accuracy in Figure \ref{spa_update}. The attack is conducted upon classifiers trained by naive, attribute robust, and SPA training.
We observed that parameter optimization could effectively enhance the SPA samples. The accuracy becomes stable after 30 steps. We also notice that the SPA-trained classifier has the smallest relative accuracy decline (7.2\%=$\frac{93.75\% - 86.97\%}{93.75\%}$ for 50 steps) than the other baselines (26.1\% for attribute robust training and 56.6\% for naive training). For efficiency, we only update 10 steps in SPA training.
\begin{figure}[t!]
  \centering
  \includegraphics[width=0.9\columnwidth]{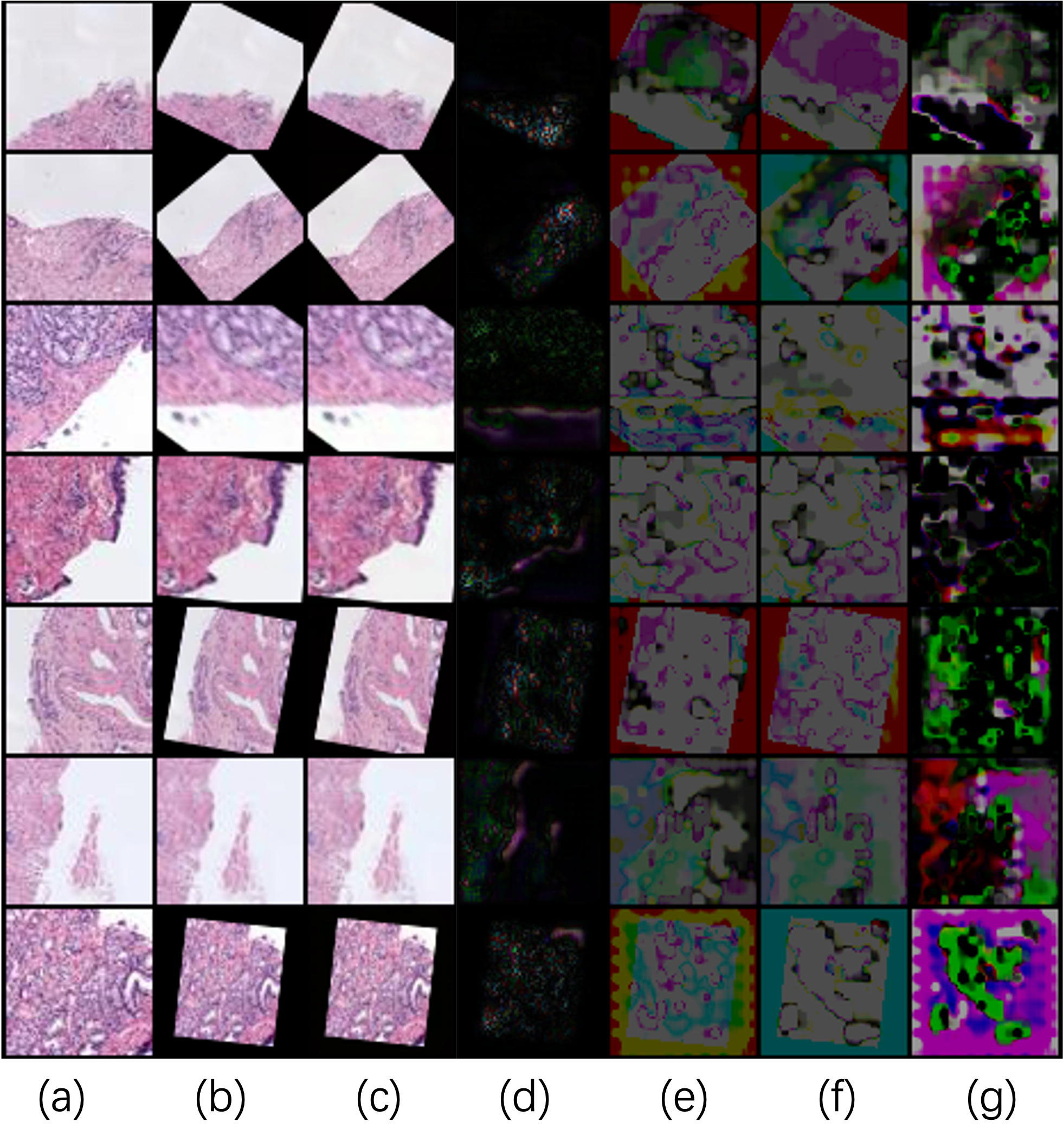}
  \caption{Semantic-preserving adversarial (SPA) examples generated with multiple attributes on the SICAPv2 dataset. Column (a) shows original images. Column (b) shows images with optimized attributes. Column (c) shows SPA images with joint perturbations. Column (d) shows the gradient of images in column (c). Column (e) and column (f) show the adversarial noises generated with different diversity variables. Column (g) shows the difference between the two noises in columns (e) and (f). Images in columns (e), (f), (g) are amplified 30 times.}
\end{figure}

\textbf{Attribute Number.}
We quantitatively explore the effect of introducing a different number of attributes for the SPA attack. We conduct experiments on the SICAPv2 dataset by attacking a classifier adversarially trained against PGD attack. Table~\ref{multi_attribute} demonstrates that as the number of attributes increases, the SPA attack can efficiently search stronger natural perturbations by optimizing in larger attribute space. 
Compared to randomly sampling attributes as in Rand-SPA, our optimization-based SPA attack fears better and can find stronger adversarial samples.

\subsection{Comparisons}
In this section, we compare our approach to existing works in defense and attack. 
We use four popular defense methods in the literature, AGAT~\cite{gokhale2021attribute}, CW~\cite{Carlini2017CandW}, PGD~\cite{madry2018towards} and L2LDA~\cite{jang2019adversarial}, as the baselines for adversarial training. 
The results on MNIST, FashionMNIST, CelebA, and SICAPv2 datasets are presented in Table~\ref{attack_defence_table}. 

First, we observe that the SPA-trained target classifier outperforms the others by a significant margin in all four datasets against jointly-perturbed samples. 
Compared to AGAT and SPA training, the two classifiers trained by CW and L2LDA cannot defend natural perturbations from SAA in most cases. 
SPA training achieves comparable performance to AGAT against natural perturbations. 
However, the SPA-trained classifier is robust against adversarial perturbations, while the AGAT-trained classifier is not. 
According to the fifth to the seventh columns in Table~\ref{attack_defence_table}, our proposed SPA training is robust against both perturbations. 

To well evaluate the performance of our proposed SPA attack, we use three existing attack methods  (SAA~\cite{joshi2019semantic}, CW, and L2LDA) and two hybrid attack methods (CW-Attr and L2LDA-Attr) as baselines. We observe that the SPA attack outperforms other natural-perturbation-based attacks, gradient-based adversarial attacks, generator-based adversarial attacks, and hybrid attacks in most cases. 
Although random attribute perturbation can help enhance adversarial samples, the SPA attack generates stronger samples thanks to parameter optimization. 
Therefore, the SPA attack is able to generate powerful jointly-perturbed samples for improving the robustness against joint perturbations. 

\section{Conclusion and Future Work}
This paper proposes a new adversarial training strategy named Semantic-Preserving Adversarial (SPA) training for enhancing robustness against joint perturbations in the attribute and pixel spaces by designing a novel attack mechanism. 
To make the classifier more robust against joint adversarial and natural perturbations, we leverage an attribute manipulator for natural perturbation and a noise generator to generate diverse adversarial noises, then optimize both attribute values and adversarial diversity variables. 
The SPA attack causes a larger performance decline under small $l_{\infty}$ norm-ball constraints compared to existing approaches. 
We extensively evaluate SPA attacks and training on four benchmarks and achieve state-of-the-art performance. Besides, we empirically demonstrate that SPA training applies to multiple types of natural perturbations and can be used with different surrogate functions for attribute manipulation. In the future, we are going to explore joint-perturbation space with a unified generator more effectively and adapt our SPA training to other robustness problems, not limited to classification.

\bibliographystyle{IEEEtran}
\bibliography{ref}
\end{document}